\documentclass{article} % For LaTeX2e
\usepackage{iclr2027_conference,times}

\usepackage{amsmath,amsfonts,bm}

\def\eqref#1{equation~\ref{#1}}
\def\1{\bm{1}}

\DeclareMathAlphabet{\mathsfit}{\encodingdefault}{\sfdefault}{m}{sl}
\SetMathAlphabet{\mathsfit}{bold}{\encodingdefault}{\sfdefault}{bx}{n}

\usepackage{hyperref}
\usepackage{url}
\usepackage{graphicx}
\usepackage{booktabs}
\usepackage{xcolor}
\usepackage{colortbl}
\usepackage{placeins}

\usepackage{amsthm}
\newtheorem{proposition}{Proposition}

\title{Contrastive On-Policy Distillation\\ via Dual-Branch Supervision}

\author{
Zhibo Hou, Fan Zhao, Yiping Li, Zhiyu An, Wan Du \\
University of California, Merced \\
\texttt{\{zhou6,fanzhao,yipingli,zan7,wdu3\}@ucmerced.edu}
}

\iclrfinalcopy % Uncomment for camera-ready version, but NOT for submission.
\begin{document}

\maketitle
\fancyhead{}

\begin{abstract}
For on-policy distillation (OPD), more capable models are a natural choice as teachers. However, higher task performance does not necessarily imply greater teachability. A controlled token intervention study suggests that even a model weaker than the student can serve as an informative contrastive reference.
Building on this observation, we propose Contrastive On-Policy Distillation (cOPD), which provides dual branch supervision by contrasting strong teacher, weak teacher, and student predictions at student generated prefixes.
Our analysis explains how this three way comparison helps select token candidates associated with better or worse reasoning outcomes under an idealized model.
We evaluate cOPD across diverse teacher student configurations spanning different training stages, parameter scales, and model families, on tasks including mathematical reasoning, instruction following, and question answering with abstention.
cOPD achieves the best or top-tier performance among the evaluated baselines across all three tasks, improving overall Pass@16 on mathematical reasoning by up to 7.51\% over the strongest baseline. Further analyses show that cOPD suppresses undesirable response patterns and suggest that it facilitates exploration.
\end{abstract}

\section{Introduction}
Distillation has rapidly emerged as a core technique for large language model (LLM) post-training. In particular, On-Policy Distillation (OPD)~\citep{agarwal2024policy} samples trajectories from the student and obtains supervision from the teacher’s next-token distributions at the resulting prefixes, providing dense, token level guidance.
% on the student’s own generation distribution and aligning training-time supervision with its inference-time generation process. 
Recent LLMs, including Qwen3~\citep{yang2025qwen3}, DeepSeek-V4~\citep{xu2026deepseek}, and GLM-5~\citep{zeng2026glm}, have incorporated OPD into their post-training pipelines for capability transfer and consolidation, highlighting its growing practical importance alongside supervised fine-tuning (SFT) and reinforcement learning (RL).

As OPD uses the teacher’s predictive distribution as a training target, using more capable models as teachers is both a natural and widely adopted design choice ~\citep{jin2026entropy, yang2025qwen3, gu2024minillm}. However, recent work shows that higher teacher accuracy does not necessarily translate into greater student improvement through OPD~\citep{li2026rethinking, fu2026rethinking}. A model’s ability to solve tasks is therefore distinct from its teachability in distilling a particular student. Although a model with lower performance may be a less reliable target for direct distillation, its distribution might still contain useful information for improving the student.

\begin{figure}[t]
    \centering
    \includegraphics[width=1\linewidth]{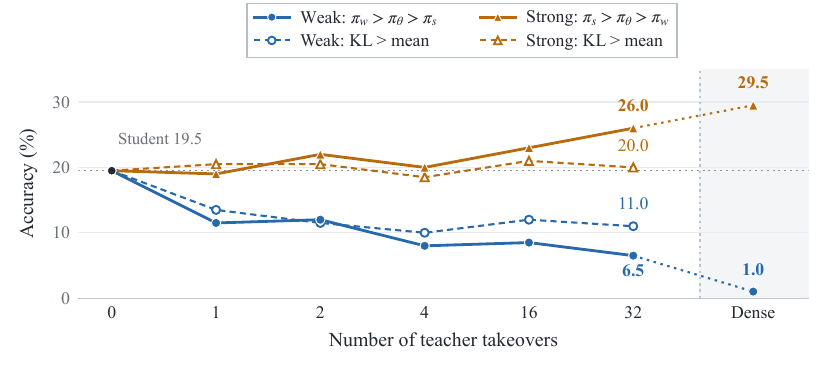}
    \caption{Teacher takeover experiment. As the takeover budget increases, the negative branch with weak teacher takeovers generally reduces student accuracy, whereas the positive branch with strong teacher takeovers generally improves it. Both effects are larger than under KL-based selection.}
    \label{fig:motivation}
\end{figure}

This key observation motivates our research question in this work: \textbf{Can a model that performs worse than the student still provide beneficial supervision?} We investigate weaker model's role as an auxiliary teacher, using its predictive distribution as a contrast to that of a primary, stronger teacher. Our hypothesis is that comparing their token level preferences can help identify both beneficial token choices to reinforce and harmful ones to suppress. A token intervention study provides initial support for this hypothesis. Figure \ref{fig:motivation} shows that takeovers guided by weak and strong teacher contrasts can improve student accuracy when favoring the strong teacher and reduce it when favoring the weak teacher. 

In this paper, we propose Contrastive On-Policy Distillation (cOPD). At student generated prefixes, cOPD selects token candidates whose probabilities under the student lie between those assigned by the strong and weak teachers. Through the teachers’ log-probability contrasts, cOPD provides dual-branch supervision, where the positive branch reinforces selected preferences favored by the strong teacher, while the negative branch suppresses those favored by the weak teacher. Our analysis explains how comparing the strong teacher, student, and weak teacher helps select token candidates associated with better or worse reasoning outcomes under an idealized model. 
We evaluate cOPD across diverse teacher student configurations on mathematical reasoning, instruction following, and question answering with abstention, and find that it achieves the best or top-tier performance among the evaluated baselines.
Further empirical analyses show that cOPD effectively uses weak teacher information to suppress undesirable response patterns, with qualitative evidence suggesting benefits for exploring additional valid responses.

Our specific contributions are as follows:
\begin{enumerate}
    \item We show through controlled token takeovers that weak teachers provide complementary information for identifying beneficial and harmful token preferences.
    \item We propose cOPD, which uses strong weak teacher contrasts to selectively reinforce beneficial token preferences and suppress harmful ones.
    \item We provide extensive empirical evidence across three tasks and diverse teacher student configurations that cOPD achieves best or near-best performance while suppressing undesirable failure modes.
\end{enumerate}
\vspace{-1em}
\section{Preliminary}
Let $x \sim  \mathcal{D}$ denote an input prompt and $y = (y_1, ...  ,y_L)$ a response, with $y_{<t} = (y_1, ..., y_{t-1})$.
We denote the student and a teacher by $\pi_\theta$ abd $\pi_T$, with their next token distributions as $\pi(\cdot|x, y_{<t})$.
% An autoregressive language model factorizes the conditional probability of the response as
% \begin{equation*}
%     \pi(y|x) = \prod^L_{t=1}\pi(y_t|x,y_{<t})
% \end{equation*}
% where $y_{<t} = (y_1, ...,y_{t-1})$ denotes the response prefix and $\pi(\cdot|x,y_{<t})$ is the next token distribution. The student policy $\pi_\theta$ is parameterized by $\theta$, and $\pi_T$ represents a generic teacher policy.
% At step $t$, an autoregressive language model predicts the next token conditioned on $x$ and the response prefix $y_{<t} = (y_1, ...,y_{t-1})$. The student policy $\pi_\theta$ parameterized by $\theta$, and a generic teacher policy $\pi_T$ provides next token distributions $\pi_\theta(\cdot|x, y_{<t})$ and $\pi_T(\cdot|x, y_{<t})$, respectively.

\paragraph{On-Policy Distillation.} OPD trains the student policy using teacher's supervision on its own generated trajectories. Given a prompt $x$ from a training set $\mathcal{D}$, the student generates a response $y \sim \pi_\theta(\cdot|x)$. At each position $t$, the teacher provides a next-token distribution $\pi_T(\cdot|x, y_{<t})$, conditioned on the prompt and the student-generated prefix. The student is updated by minimizing a token-level divergence between the teacher and student distributions:
\begin{equation*}
    \mathcal{L}_{OPD}(\theta)=\mathbb{E}_{x\sim\mathcal{D},y\sim\pi_\theta(\cdot|x)}[\frac{1}{L}\sum^{L}_{t=1}D(\pi_T(\cdot|x, y_{<t}), \pi_\theta(\cdot|x, y_{<t}))]
\end{equation*}
where $D$ is a divergence measure and the sampled trajectories are treated as fixed during each update.
\paragraph{Distillation objectives.} Common choices for the token level divergence include forward KL, reverse KL, and Jensen-Shannon divergence (JSD) \citep{song2026survey}. Let $p(v)=\pi_T(v|x,y_{<t})$ and $q(v)=\pi_\theta(v|x,y_{<t})$, where $v\in\mathcal{V}$ and $\mathcal{V}$ is the full vocabulary. The Kullback-Leibler (KL) \citep{kullback1951information} divergence between two distribution $r(v)$ and $s(v)$ is defined as:
\begin{equation*}
    D_{KL}(r||s) = \sum_{v\in\mathcal{V}}r(v)\log\frac{r(v)}{s(v)}
\end{equation*}

\textit{Forward KL}, $D_{FKL}(p,q) = D_{KL}(p||q)$, encourages mode-covering behavior by penalizing insufficient student probability on tokens supported by the teacher. 

\textit{Reverse KL}, $D_{RKL}(p,q) = D_{KL}(q||p)$, encourages mode-seeking behavior by penalizing student probability mass on tokens unlikely under the teacher. 

\textit{Jensen-Shannon divergence}, $D_{JSD}(p,q) = \frac{1}{2}D_{KL}(p||m) + \frac{1}{2}D_{KL}(q||m)$, where $m=\frac{1}{2}(p + q)$, provides a symmetric, bounded alternative by comparing both distributions to their mixture.

Most OPD implementations approximate the divergence using a top-$k$ subset of the vocabulary \citep{jin2026entropy, li2026rethinking}.

\section{Method}
\subsection{Can a Weak Teacher Help?}
OPD uses teacher supervision at student generated prefixes to reshape the student’s next token predictions. We design a controlled token intervention experiment to investigate whether a weak teacher can serve as a contrastive reference for identifying critical positions and token distributions that affect the correctness of the student’s final answer. The experiment uses temporary teacher takeovers as a simplified behavioral proxy for the effects of OPD supervision.

We use DeepSeek-R1-Distill-Qwen-1.5B ~\citep{deepseekai2025deepseekr1incentivizingreasoningcapability} as the student $\pi_\theta$, Qwen-2.5-Math-1.5B ~\citep{yang2024qwen25mathtechnicalreportmathematical} as the weak teacher $\pi_w$, and JustRL-DeepSeek-1.5B~\citep{he2025justrl} as the strong teacher $\pi_s$. We select a fixed set of 50 problems from the DAPO-Math-17k dataset~\citep{yu2025dapoopensourcellmreinforcement} for which the weak teacher $\text{pass}@8=0$. Each intervention setting is evaluated on this set with four random seeds. Given the current prefix $h_t=(x, y_{<t})$, the three models assign probabilities to the candidate tokens $v\in\operatorname{Top}_{8}(\pi(\cdot|h_t))$. We define two takeover conditions as:
\begin{equation*}
\label{eq:takeover_conditions}
\begin{aligned}
\text{Weak teacher takeover:}\quad
& \pi_{\mathrm{w}}(v \mid h_t)
> \pi_\theta(v \mid h_t)
> \pi_{\mathrm{s}}(v \mid h_t), \\
\text{Strong teacher takeover:}\quad
& \pi_{\mathrm{s}}(v \mid h_t)
> \pi_\theta(v \mid h_t)
> \pi_{\mathrm{w}}(v \mid h_t),
\end{aligned}
\end{equation*}
At the first eligible position during student generation, the corresponding teacher takeover and generates four consecutive tokens. 
% We use this short, continuous block for stability when inference returns to the student. 
The student then resumes generation until the next eligible position. This procedure repeats until the takeover budget is reached, and then student completes the response. We evaluate budget of $n\in\{1, 2, 4, 16, 32, \infty\}$ takeovers. We also evaluate a separate control that triggers takeovers where KL divergence exceeds the mean KL divergence.

As shown in Figure \ref{fig:motivation}, contrastive based selection produces an overall increase in student accuracy with strong teacher takeovers and decrease with weak teacher takeovers as the budget grows. These effects are more significant than takeovers under KL based selection suggesting that weak teacher may help to provide additional information for identifying takeover positions and distributions that improve or degrade student accuracy. 
This motivates incorporating these weak strong teacher contrasts into OPD supervision to reinforce beneficial distributions and suppress harmful ones.

\subsection{contrastive On-Policy Distillation}
Motivated by these findings, we propose Contrastive On-Policy Distillation (cOPD), which uses three way probability comparisons among the strong teacher, student, and weak teacher to construct token level supervision at student generated prefixes.
\subsubsection{Contrastive token mask}
We first define a token level mask that selects token candidates for which the student probability lies between the weak and strong teacher probabilities, as in the motivation experiment. To exclude near ties, we require both probability gaps to exceed a margin $\delta$.
% We first define a token level mask based on the probability orderings examined in the intervention study. A candidate is selected when the reference student's probability lies between those of the strong and weak teachers and both adjacent gaps exceed a margin

For a student generated prefix $h_t=(x,y_{<t})$, we have $\pi_s$, $\pi_w$, and $\pi_{ref}$ as the strong teacher, weak teacher, and reference student policy, respectively, where $v\in\operatorname{Top}_{16}(\pi_{ref}(\cdot|h_t))$. We define two binary masks corresponding to the positive and negative branches:
\begin{equation}
\begin{split}
    m_v^+=\mathbf{1}[\pi_s(v|h_t) - \pi_{ref}(v|h_t)>\delta \wedge \pi_{ref}(v|h_t) - \pi_w(v|h_t) > \delta], \\
    m_v^-=\mathbf{1}[\pi_{w}(v|h_t) - \pi_{ref}(v|h_t) > \delta \wedge \pi_{ref}(v|h_t) - \pi_{s}(v|h_t)>\delta],
\end{split}
\end{equation}
where $\mathbf{1}[\cdot]$ is the indicator function. We keep $\pi_{ref}$ fixed within each epoch and set $\pi_{ref} \leftarrow \pi_\theta$ at the end of the epoch. The positive branch selects tokens favored by the strong teacher relative to the reference student and disfavored by the weak teacher. The negative branch selects tokens with the reverse ordering. Tokens satisfying neither condition fall into the abstain case. The overall selection mask is
\begin{equation}
\label{eq:selection_mask}
    m_v=m_v^+ + m_v^-.
\end{equation}
\subsubsection{Contrastive objective}
We first quantify the relative preference of the strong and weak teachers through their log probability contrast:
\begin{equation}
\label{eq:teacher_contrast}
    \Delta_v= \log\pi_s(v) - \log\pi_w(v)=\log\frac{\pi_s(v)}{\pi_w(v)}.
\end{equation}
A positive $\Delta_v$ indicates that the strong teacher prefer token $v$ over the weak teacher, while a negative value indicates the reverse. Its magnitude reflects the difference in their relative preferences.

Let $q_\theta(v) = \pi_\theta(v|h_t)$ be the current student probability. We combine the teacher contrast with the token mask to define the per-position loss:
\begin{equation}
    \ell_{cOPD}(\theta; h_t) = -\sum_{v\in\operatorname{Top}_{16}(\pi_{ref})}m_vq_\theta(v)\tanh{\frac{\Delta_v}{\tau}}.
\end{equation}
where the temperature $\tau>0$ controls the scale of the contrast. Positive branch tokens have $\Delta_v >0$ and encourage student probability mass on these tokens, and negative branch tokens have $\Delta_v<0$ and penalize student probability mass. Abstaining tokens contribute no term to the loss.

\subsection{Analysis of contrastive supervision}

To understand how a weak teacher can provide complementary information for selecting training signals, we present a theoretical analysis.

Prior work analyzes token level teacher guidance through downstream success probabilities, distinguishing beneficial, neutral, and detrimental contributions~\citep{armandpour2026unmasking}.
For a fixed student generated prefix $h$, we model the idealized next token distribution of an autoregressive language policy as the following mixture:
\begin{equation}
    p_i=\alpha H_i + \beta_i G + \gamma_i B ,\quad i\in\{s,ref,w\},
\end{equation}
where $p_s$, $p_{ref}$, $p_w$ correspond to the strong teacher, reference student, and weak teacher, and $\alpha + \beta_i + \gamma_i=1$. The normalized distribution $G$ and $B$ are shared across models and represent good and bad reasoning candidates, that under a fixed student continuation, their correctness probabilities are respectively above and below the reference student's expected correctness at $h$. The task neutral components $H_i$ may differ across models, such as preferring “but” over “however” when the two expressions are semantically equivalent in context. We assume that the supports of $G$, $B$, and the combined neutral region are mutually disjoint, with $0<\alpha<1$ and $0<\beta_w<\beta_{ref}<\beta_s<1-\alpha$.
% (strong teacher distribution have more good reasoning candidates over weak teacher).

\begin{proposition}[Contrastive selection of reasoning preferences]
\label{prop:contrastive-selection}
Under the shared component model, token candidates in the reasoning components satisfy
\[
\begin{aligned}
v\in\operatorname{supp}(G):\quad&
p_{\mathrm{s}}(v)>p_{ref}(v)>p_{\mathrm{w}}(v),
\qquad m_v^{-}=0,\\
v\in\operatorname{supp}(B):\quad&
p_{\mathrm{w}}(v)>p_{{ref}}(v)>p_{\mathrm{s}}(v),
\qquad m_v^{+}=0.
\end{aligned}
\]
% Consequently, the cOPD loss encourages probability mass on
% selected good reasoning candidates and penalizes mass on
% selected bad reasoning candidates.
Within the reasoning components, $m_v^{+}$ and $m_v^{-}$
serve as selective proxies for good and bad reasoning choices,
respectively. The cOPD loss uses these proxies to encourage
probability mass on selected good reasoning candidates and
penalize mass on selected bad reasoning candidates.
\end{proposition}

\renewcommand{\qedsymbol}{}
\begin{proof}[Proof]
For $v\in\operatorname{supp}(G)$, the disjoint-support assumption gives
$p_{\mathrm{s}}(v)-p_{ref}(v)
=(\beta_{\mathrm{s}}-\beta_{ref})G(v)$ and
$p_{ref}(v)-p_{\mathrm{w}}(v)
=(\beta_{ref}-\beta_{\mathrm{w}})G(v)$.
Both gaps are positive, yielding
$p_{\mathrm{s}}(v)>p_{ref}(v)>p_{\mathrm{w}}(v)$,
while the negative branch remains inactive.
Since $\gamma_i=1-\alpha-\beta_i$, the same argument applies
to $B$ with the teacher ordering reversed, yielding
$p_{\mathrm{w}}(v)>p_{ref}(v)>p_{\mathrm{s}}(v)$
and $m_v^{+}=0$.
The teacher log-ratio is positive on $G$ and negative on $B$,
giving the corresponding reinforcement and suppression
directions in the loss.
\end{proof}

The probability margin $\delta$ further restricts which candidates are selected without changing these directional conclusions.
The analysis illustrates how a teacher that assigns less probability mass to good reasoning can still provide an informative contrastive reference. cOPD uses this additional reference to select supervision from the relative predictions of the strong and weak teachers.
\section{Experiments}
% \color{red}placeholder for research questions?\color{black}

\begin{table}[t]
    \centering
    \caption{Results on mathematical reasoning (\%). Avg.\ is the overall
    Pass@16 across all problems pooled from AIME24/AIME25/AIME26/AMC23.
    Bold and underlined scores indicate the best and second best results.}
    \label{tab:math-aime26}
    \vspace{\baselineskip}
    \small
    
    \setlength{\tabcolsep}{3pt}
    \renewcommand{\arraystretch}{1.00}
    \setlength{\aboverulesep}{1pt}
    \setlength{\belowrulesep}{1pt}
    \begin{tabular*}{\linewidth}{@{\extracolsep{\fill}}l*{6}{c}@{}}
        \toprule
        \textbf{Method}
        & \shortstack[c]{AIME24\\Pass@16 $\uparrow$}
        & \shortstack[c]{AIME25\\Pass@16 $\uparrow$}
        & \shortstack[c]{AIME26\\Pass@16 $\uparrow$}
        & \shortstack[c]{AMC23\\Pass@16 $\uparrow$}
        & \shortstack[c]{Avg.\\Pass@16 $\uparrow$}
        & \shortstack[c]{Omni-Math\\Pass@1 $\uparrow$} \\

        \midrule
        \rowcolor{black!6}
        \multicolumn{7}{c}{\makebox[0pt][c]{\textit{JustRL / Qwen2.5-Math / DeepSeek-R1-Distill}}} \\
        \midrule
        Base & 70.00 & 46.67 & 40.00 & 87.95 & 69.36 & 16.84 \\
        Strong teacher & 76.67 & 60.00 & 40.00 & 93.98 & 75.72 & 28.18 \\
        Weak teacher & 23.33 & 13.33 & 10.00 & 66.27 & 39.88 & 10.35 \\
        \midrule
        FKL & \underline{76.67} & \underline{53.33} & \underline{43.33} & 91.57 & 73.99 & 12.19 \\
        RKL & 73.33 & 46.67 & 40.00 & \underline{93.98} & 72.83 & 15.14 \\
        JSD & \underline{76.67} & \textbf{56.67} & \underline{43.33} & 92.77 & \underline{75.14} & 14.14 \\
        MAD-OPD & 73.33 & 50.00 & 40.00 & 87.95 & 70.52 & 16.66 \\
        W2S-OPD & \underline{76.67} & 46.67 & 36.67 & \underline{93.98} & 72.83 & \textbf{25.49} \\
        \midrule
        \textbf{cOPD (ours)} & \textbf{80.00} & \textbf{56.67} & \textbf{46.67} & \textbf{95.18} & \textbf{77.46} & \underline{23.82} \\
        \midrule
        \rowcolor{black!6}
        \multicolumn{7}{c}{\makebox[0pt][c]{\textit{Qwen3-4B / 0.6B / 1.7B}}} \\
        \midrule
        Base & 33.33 & 30.00 & 36.67 & 74.70 & 53.18 & 11.63 \\
        Strong teacher & 46.67 & 46.67 & 53.33 & 83.13 & 65.32 & 20.60 \\
        Weak teacher & 13.33 & 13.33 & 10.00 & 56.63 & 33.53 & 11.34 \\
        \midrule
        FKL & \underline{46.67} & 33.33 & 26.67 & \underline{78.31} & 56.07 & \underline{16.13} \\
        RKL & 36.67 & \underline{36.67} & 20.00 & 77.11 & 53.18 & 15.92 \\
        JSD & 43.33 & 30.00 & \underline{30.00} & \underline{78.31} & 55.49 & 15.77 \\
        MAD-OPD & 36.67 & 23.33 & 16.67 & 75.90 & 49.71 & 15.95 \\
        W2S-OPD & 40.00 & \underline{36.67} & \underline{30.00} & \textbf{81.93} & \underline{57.80} & 16.02 \\
        \midrule
        \textbf{cOPD (ours)} & \textbf{63.33} & \textbf{50.00} & \textbf{36.67} & \textbf{81.93} & \textbf{65.32} & \textbf{16.41} \\
        \bottomrule
    \end{tabular*}
\end{table}
We evaluate whether cOPD improves On-Policy Distillation by using a weak teacher as a contrastive reference. Our experiments span mathematical reasoning, instruction following, and question answering with abstention. The teacher student configurations cover different post-training backgrounds, parameter scales, and model families.

\subsection{Experiment setup}
We use three model configurations, listed below in the order of strong teacher, weak teacher, and student:
\begin{enumerate}
    \item JustRL-DeepSeek-1.5B~\citep{he2025justrl}, Qwen2.5-Math-1.5B~\citep{yang2024qwen25mathtechnicalreportmathematical}, and DeepSeek-R1-Distill-Qwen-1.5B~\citep{deepseekai2025deepseekr1incentivizingreasoningcapability}. These models share Qwen2.5-1.5B as base model but differ in post-training.
    \item Qwen3-4B, Qwen3-0.6B, and Qwen3-1.7B~\citep{yang2025qwen3}. These models belong to the same family but differ in parameter scale.
    \item Llama-3.1-8B-Instruct, Llama-3.2-1B-Instruct, and Llama-3.2-3B-Instruct. These models span different releases and parameter scales within the Llama family.
\end{enumerate}
% We run evaluation on three separate task: math reasoning, instruction following, and question answering with abstention. The corresponding training datasets are filtered subsets of DAPO-Math-17k\citep{yu2025dapoopensourcellmreinforcement}, IFEval\citep{pyatkin2025generalizing}, and SQuAD2.0\citep{rajpurkar-etal-2018-know}, respectively. We retain data points that the weak teacher produces no correct response (Pass@8 = 0).
The model configurations used for each task, together with the corresponding training and evaluation settings, are described in the following subsections.

We compare cOPD against six baselines: the initial student model (Base), OPD with forward KL (FKL), reverse KL (RKL), or Jensen--Shannon divergence (JSD), multi teacher debate OPD (MAD-OPD)~\citep{wang2026madopdbreakingceilingonpolicy}, and weak to strong multi treacher OPD (W2S-OPD)~\citep{yu2026weaktostrongonpolicydistillation}. All models are run in non-thinking mode with detailed training hyperparameters provided in Appendix \ref{app:implementation}.

\subsection{Mathematical Reasoning}
% Preamble: \usepackage{booktabs}, \usepackage{xcolor}, \usepackage{colortbl}
% If xcolor is already loaded by the template, only add colortbl.

\paragraph{Setup and evaluation.}
We use DAPO-Math-17k~\citep{yu2025dapoopensourcellmreinforcement} for training with the three model configurations. We exclude the Llama configuration because we do not observe a clear performance gap between the teachers and the student on mathematical reasoning. For each configuration, we retain prompts for which the weak teacher produces no correct response and the strong teacher produces at least one correct response in eight samples. This filtering results in 6,016, 7,488 training prompts for the JustRL and Qwen3 configurations. We evaluate on AIME24~\citep{aime24}, AIME25~\citep{aime25}, AIME26~\citep{aime26}, and AMC23 using Pass@16, and on Omni-Math~\citep{gao2025omni} using Pass@1.

\paragraph{Results.}
We report mathematical reasoning results in Table \ref{tab:math-aime26}. cOPD achieve the highest overall Pass$@16$ among all distillation methods in both configurations, reaching $77.46\%$ with JustRL and $65.32\%$ with Qwen3. These results outperform the strongest baselines by up to 7.51 percentage points. More importantly, cOPD surpasses the strong teacher in the JustRL configuration and matches the strong teacher in the Qwen3 configuration. On Omni-Math, cOPD achieves the best student performance with Qwen3 and ranks second to W2S-OPD with JustRL.

We further analyze truncation as a failure mode in the JustRL configuration in Figure \ref{fig:failure_analysis}(a). The weak teacher exhibits frequent truncation errors and low accuracy, whereas the strong teacher achieves higher accuracy with fewer truncation errors. Strong-teacher-only OPD transfers this favorable behavior, reducing the student's truncation errors from 321 to 54–65 while improving accuracy.

Incorporating the weak teacher, however, does not consistently preserve these benefits. MAD-OPD has 318 truncation errors with no accuracy gain over the initial student. This suggests that a weak teacher whose responses are frequently truncated may provide limited useful feedback during debate, limiting student improvement. W2S-OPD matches the strong teacher's accuracy but retains more truncation errors than single-teacher OPD, suggesting that their incorporation of weak teacher may limit the suppression of truncation failures.

cOPD reduces truncation errors to 56 while achieving the highest accuracy, even surpassing the strong teacher. These results suggest that cOPD can exploit the weak teacher's complementary information to improve performance while suppressing its failure modes.

\subsection{Instruction Following}
\begin{table}[t]
    \centering
    \caption{
        Results on instruction following (IF) and question answering
        with abstention (QA w/ Abstention). All scores are percentages, and
        Abst.\ represents the correct abstention rate.
        Bold indicates the best student result in each column, and underline indicates the second best.
    }
    \label{tab:if-qa}
    \vspace{4pt}
    \small
    \setlength{\tabcolsep}{3pt}
    \renewcommand{\arraystretch}{1.08}

    \begin{tabular*}{\linewidth}{@{\extracolsep{\fill}}l*{8}{c}@{}}
        \toprule
        & \multicolumn{4}{c}{\makebox[0pt][c]{\shortstack[l]{\textit{
            Qwen3-4B/ 0.6B/ 1.7B}}}}
        & \multicolumn{4}{c}{\makebox[0pt][c]{\shortstack[l]{\textit{
            Llama-3.1-8B/ 3.2-1B/ 3.2-3B}}}} \\
        \cmidrule(lr){2-5}
        \cmidrule(lr){6-9}
        & IF & \multicolumn{3}{c}{QA w/ Abstention}
        & IF & \multicolumn{3}{c}{QA w/ Abstention} \\
        \cmidrule(lr){2-2}
        \cmidrule(lr){3-5}
        \cmidrule(lr){6-6}
        \cmidrule(lr){7-9}
        Method
        & Acc.$\uparrow$ & EM$\uparrow$ & F1$\uparrow$ & Abst.$\uparrow$
        & Acc.$\uparrow$ & EM$\uparrow$ & F1$\uparrow$ & Abst.$\uparrow$ \\
        \midrule
        Base
        & 20.36 & 60.14 & 65.64 & 55.68
        & 17.81 & 54.20 & 60.82 & 46.24 \\

        Strong teacher
        & 27.13 & 68.35 & 74.39 & 62.32
        & 18.74 & 67.07 & 71.75 & 69.50 \\

        Weak teacher
        & 18.33 & 35.01 & 39.80 & 30.63
        & 11.69 & 16.63 & 21.23 & 8.53 \\
        \midrule

        % Base
        % & 20.37 & 60.14 & 65.64 & 55.68
        % & 17.82 & 54.20 & 60.82 & 46.24 \\

        FKL
        & 23.20 & \underline{71.43} & \underline{76.49} & \underline{69.86}
        & 19.35 & 65.71 & 70.05 & 68.23 \\

        RKL
        & \underline{23.46} & 67.75 & 73.33 & 63.13
        & 19.11 & 66.43 & 70.65 & 68.73 \\

        JSD
        & 22.48 & 70.04 & 75.40 & 68.33
        & 18.66 & \underline{67.36} & \underline{71.75} & \underline{70.58} \\

        MAD-OPD
        & 20.10 & 63.18 & 69.97 & 57.78
        & 16.07 & 48.75 & 51.10 & 64.44 \\

        W2S-OPD
        & 22.81 & 66.17 & 71.28 & 59.60
        & \underline{19.77} & 3.40$^{\dagger}$ & 4.01$^{\dagger}$ & 1.95$^{\dagger}$ \\
        \midrule

        \textbf{cOPD (ours)}
        & \textbf{24.26} & \textbf{72.09}
        & \textbf{77.31} & \textbf{71.59}
        & \textbf{21.73} & \textbf{68.97}
        & \textbf{73.17} & \textbf{73.04} \\
        \bottomrule
    \end{tabular*}

    \vspace{3pt}
    \begin{minipage}{\linewidth}
        \footnotesize
        \raggedright
        $^{\dagger}$ For Llama QA, W2S-OPD exhibits a catastrophic formatting error
        that causes verifier failures. Ignoring this error yields
        EM, F1, and correct abstention rate scores of
        34.69, 41.27, and 13.29.
    \end{minipage}
\end{table}
% \FloatBarrier
% \vspace{-10pt}
\begin{figure}[h]
    \centering
    \includegraphics[width=\linewidth]{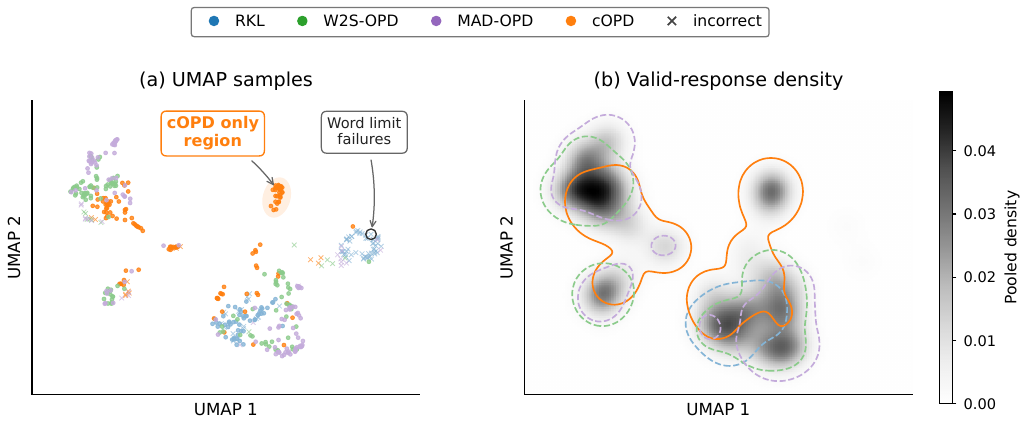}
    \caption{
        Instruction-following response distributions and constraint
        satisfaction in a shared UMAP space, where each point represents a response and crosses indicate violation. 
        cOPD produces valid responses in additional regions while
        largely avoiding the prominent word-limit failure region.
    }
    \vspace{-0.5em}
    \label{fig:if_umap}
    \vspace{-0.5em}

\end{figure}
% \vspace{-0.5em}

\paragraph{Setup and evaluation.}
We use the multi constraint IF-RLVR~\citep{pyatkin2025generalizing} dataset, with up to five verifiable constraints per prompt. A fixed set of 7,000 prompts is held out for testing. For each model configuration, we retain training prompts for which the weak teacher produce no correct response, and the strong teacher produces at least one correct response in eight samples, yielding 6,892 and 6,903 training prompts for Qwen3 and Llama.

\paragraph{Results.}
We report evaluation results in Table \ref{tab:if-qa}. cOPD consistently outperforms all distillation baselines in both model configurations, achieving $24.26\%$ accuracy with Qwen3, and $21.73\%$ with Llama. These results exceed the strongest baseline in each configuration, RKL for Qwen3 and W2S-OPD for llama, by $0.80$ and $1.96$ percentage points, respectively. Moreover, on Llama, cOPD surpasses the strong teacher by $3.00$ percentage points. 
% The comparison across methods suggests that the benefit of an additional weak teacher depends on how its information is incorporated. For MAD-OPD, the weak teacher provides limited useful information during debate while still serving as an imitation target, which may contribute to the observed performance degradation relative to the base student model on both Qwen3 and Llama.

Beyond the accuracy results, we further visualized responses generated using identical prompts in a shared UMAP space in Figure \ref{fig:if_umap}. Our method covers the major regions containing valid baseline responses. It also reach a distinct cluster of correct responses absent from the baseline samples, demonstrating its ability to explore additional regions while satisfying all instruction constraints. Meanwhile, cOPD largely
avoids the prominent word limit failure region, confirming our hypothesis that cOPD can suppress harmful response patterns. We provide illustrative response excerpts in Appendix \ref{app:if_examples}.

\subsection{QA with Abstention}
\paragraph{Setup and evaluation.} We use SQuAD2.0~\citep{rajpurkar2018know}, which contains both answerable and unanswerable questions. Given a context and a question, the model is instructed to answer based on the context and abstain when the context does not support an answer. From the training split, we retain prompts for which the weak teacher produces no correct response and the strong teacher produces at least one correct response in eight samples. This filtering yields 8,080 and 16,359 training prompts for Qwen3 and Llama, with unanswerable questions accounting for $26.51\%$ and $49.91\%$ of the training subsets. We evaluate on the full official development split of 11,873 questions.

\paragraph{Results.} We report QA with abstention results in Table \ref{tab:if-qa}. cOPD outperforms all distillation baselines and even the strong teacher on EM, F1, and correct abstention rate in both configurations. cOPD achieves a correct abstention rate of 71.59\% on Qwen3, exceeding the strongest baseline by 1.73 percentage points and on Llama, it reaches 73.04\%, a gain of 2.46 percentage points over the strongest baseline. 
% These results are consistent with our motivation that a weak teacher can provide complementary information for distillation. By using the weak teacher as a contrastive reference, cOPD can reduce the influence of less useful or misleading preferences in the strong teacher's distribution, helping the student improve beyond direct imitation of the strong teacher.

\begin{figure}[t]
    \centering
    \includegraphics[width=\textwidth]{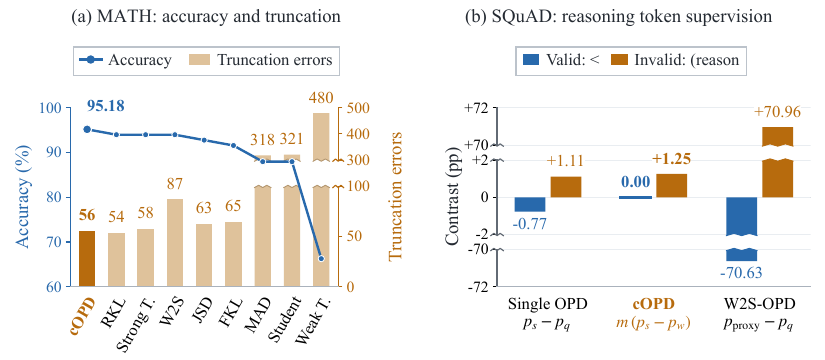}
    \caption{\textbf{Failure mode analysis.}
    (a) Mathematical reasoning accuracy (Pass@16) and truncation error counts. cOPD suppresses truncation failures while surpassing the strong teacher in accuracy.
    (b) Opening-token supervision for valid and invalid reasoning formats. W2S-OPD excessively amplifies the invalid token's relative weight, whereas cOPD's corresponding gap is negligible.
}
    \label{fig:failure_analysis}
\end{figure}

We further investigate the catastrophic degradation of W2S-OPD on SQuAD with the Llama configuration. We observe that the trained model incorrectly starts its reasoning with \texttt{(reason} instead of \texttt{<}, violating the required \texttt{<reasoning>} format.
To examine this behavior, we sampled 128 training prompts and compute the initial student's and both teachers' distributions at the first response position. Figure~\ref{fig:failure_analysis}(b) compares the teacher student probability gaps for both tokens between OPD's single teacher, W2S-OPD's proxy teacher, and the student. For cOPD, we report the masked probability contrast $m_v(p_s - p_w)$ from Equation \ref{eq:selection_mask} and \ref{eq:teacher_contrast} as a diagnostic supervision coefficient.

W2S-OPD excessively amplifies the relative weight of the invalid token \texttt{(reason}, raising its average probability from 1.25\% under the strong teacher to 71.10\% under the proxy teacher. This distorted target provides an explanation for the observed formatting failures.

In comparison, cOPD uses a margin gated mask to selectively apply contrastive supervision. Its estimated supervision coefficient for the invalid token remains small, without the extreme amplification observed in W2S-OPD. This results suggest that cOPD can use weak teacher information while remain robust to failures modes.

\subsection{Discussion of Results}
Our empirical evidence can answer the motivating research question affirmatively: \textbf{A model that performs worse than the student can provide beneficial supervision when used as a contrastive reference alongside a stronger teacher.} Across mathematical reasoning, instruction following, and question answering with abstention, cOPD achieves the best or tied-best performance among the evaluated OPD methods. These gains span different model families, parameter scales, and post training backgrounds, supporting the value of weak-teacher contrasts across the evaluated settings.

The benefits also extend to suppressing undesirable response patterns. In mathematical reasoning, cOPD achieves the highest accuracy in the analyzed configuration while substantially reducing truncation errors (Figure \ref{fig:failure_analysis}(a)). In instruction following, it largely avoids the prominent failure region associated with word limit violations (Figure \ref{fig:if_umap}). On SQuAD2.0, it avoids the catastrophic formatting failure observed with W2S-OPD in the Llama configuration (Figure \ref{fig:failure_analysis}(b)). Together, these results indicate that cOPD can use weak teacher information as an counterexample to avoid weak teacher's failure modes.

Additionally, the instruction-following visualization suggests a potential benefit for exploration, that cOPD reaches a distinct region of satisfying responses absent from the baselines (Figure \ref{fig:if_umap}). Ablations in Appendix \ref{app:ablation} show the contribution of both supervision branches and dataset selection.
\section{Related Works}
\vspace{-0.5em}
\subsection{On-Policy Distillation}
\vspace{-0.5em}
OPD trains students on their own generated trajectories using token-level teacher supervision, reducing the distribution mismatch between training and inference~\citep{song2026survey, agarwal2024policy, kim2016sequence}. Following studies have refined distillation objectives to account for teacher uncertainty~\citep{jin2026entropy, li2026rethinkingr} and examined how teacher–student compatibility affects knowledge transfer~\citep{niu2026breaking, wang2026not, liu2026teacher}. The reliance of standard OPD on a single teacher motivates exploring complementary supervision from multiple models.
\vspace{-0.5em}
\subsection{Multi-Teacher OPD}
\vspace{-0.5em}
Multi-teacher OPD methods broadly follow two approaches, teacher routing and teacher aggregation. Routing based methods keep single teacher supervision at the question level by assigning each student generated trajectory to a domain specific teacher~\citep{ma2026mopd, gao2026open}. Whereas aggregation based methods combine supervision from multiple teachers for the same trajectory~\citep{wang2026madopdbreakingceilingonpolicy}. Our work falls within the broader scope of multi teacher OPD, while focusing on a different question: how models weaker than the student can contribute useful supervision.
\vspace{-0.5em}
\subsection{Weaker Model Supervision}
\vspace{-0.5em}

Prior work has demonstrated the value of weak supervision in other learning settings, where stronger models are trained on labels generated by weaker models~\citep{burns2023weak, yuan2026incentivizing}, as well as rejected responses used as negative examples in preference learning ~\citep{wang2026drift}.

Direct-OPD~\citep{feng2026weak} uses the log probability contrast between a weak model's pre and post RL checkpoints as token level supervision for stronger students. W2S-OPD~\citep{yu2026weaktostrongonpolicydistillation} constructs a proxy teacher by adding the logit contrast between positive and negative models to the student's logits. Our method supports teacher combinations beyond paired pre and post RL checkpoints, and cOPD does not construct a proxy teacher. Instead, it uses the reference student as the anchor for three way probability comparisons to select token-level supervision.
\section{Conclusion}
In this work, we investigated whether a model that performs worse than the student can still provide beneficial supervision. Our findings support an affirmative answer that a weak teacher can serve as an informative contrastive reference alongside a stronger teacher. Building on this insight, cOPD selectively reinforces beneficial token preferences and suppresses harmful ones through dual-branch supervision. Experiments across three tasks and diverse model configurations support its effectiveness in improving student performance while suppressing undesirable response patterns. Further evidence suggests that weak teacher contrasts may encourage exploration of unique, valid responses.

We believe standard KL-based OPD does not explicitly verify whether teacher preferences improve task outcomes, so matching the teacher distribution may not yield the most beneficial training direction. This motivates selecting useful signals from raw teacher supervision. We view cOPD as a step toward this goal, using weak-teacher contrasts as complementary information to identify which token preferences to reinforce or suppress. Future work could extend cOPD to multiple teachers, combining guidance from strong teachers with complementary contrasts from diverse weak teachers to identify and suppress a broader range of failure modes.

% \section{Preparing PostScript or PDF files}

\section*{AI use statement}
In this work, we used generative AI tools to assist with implementing our proposed method. We did not use generative AI tools for research ideation. The authors manually reviewed the AI-assisted implementation for consistency with the intended method and checked its behavior in our experiments.

We also used generative AI tools to aid and polish writing. This included suggestions for preliminary drafts, alternative word choices, and sentence formulations to express our ideas more clearly, as well as grammar checks. We treated AI-suggested drafts as references and rewrote the manuscript sentence by sentence. The authors determined the final wording.

We reviewed all AI assisted work and take responsibility for the final content of this work, including text, claims or artifacts produced with the aid of generative AI.

% In this work, we used generated AI tools to assist with the implementation of our proposed method and to suggest preliminary drafts, alternative word choices, and sentences to better express our ideas, as well as to check grammar. We manually reviewed all AI assisted implementation for consistency with our proposed method and checked its behavior in our experiment. We take responsibility for the final content of this work, including text, claims or artifacts produced with the aid of generative AI.

% \subsection*{Ethics statement}

% (This section is \textbf{recommended} and does not count toward the page limit.)

% If authors feel that their paper submission raises questions regarding the Code
% of Ethics, they are encouraged to include a paragraph of Ethics Statement (at
% the end of the main text before references) to address potential concerns where
% appropriate. Topics include, but are not limited to, studies that involve human
% subjects, practices to data set releases, potentially harmful insights,
% methodologies and applications, potential conflicts of interest and sponsorship,
% discrimination/bias/fairness concerns, privacy and security issues, legal
% compliance, and research integrity issues (e.g., IRB, documentation, research
% ethics). This statement should not be more than 1 page.

\section*{Reproducibility statement}
We use publicly available Large Language Models and dataset for training and evaluation, and we have included all training hyperparameters in Appendix \ref{app:implementation}. We will release our implementation code upon acceptance.



\bibliography{iclr2027_conference}

@inproceedings{agarwal2024policy,
  title={On-policy distillation of language models: Learning from self-generated mistakes},
  author={Agarwal, Rishabh and Vieillard, Nino and Zhou, Yongchao and Stanczyk, Piotr and Ramos Garea, Sabela and Geist, Matthieu and Bachem, Olivier},
  booktitle={International Conference on Learning Representations},
  volume={2024},
  pages={21246--21263},
  year={2024}
}

@article{yang2025qwen3,
  title={Qwen3 technical report},
  author={Yang, An and Li, Anfeng and Yang, Baosong and Zhang, Beichen and Hui, Binyuan and Zheng, Bo and Yu, Bowen and Gao, Chang and Huang, Chengen and Lv, Chenxu and others},
  journal={arXiv preprint arXiv:2505.09388},
  year={2025}
}

@article{zeng2026glm,
  title={Glm-5: from vibe coding to agentic engineering},
  author={Zeng, Aohan and Lv, Xin and Hou, Zhenyu and Du, Zhengxiao and Zheng, Qinkai and Chen, Bin and Yin, Da and Ge, Chendi and Huang, Chenghua and Xie, Chengxing and others},
  journal={arXiv preprint arXiv:2602.15763},
  year={2026}
}

@article{xu2026deepseek,
  title={Deepseek-v4: Towards highly efficient million-token context intelligence},
  author={Xu, Anyi and Lin, Bangcai and Xue, Bing and Wang, Bingxuan and Xu, Bingzheng and Wu, Bochao and Zhang, Bowei and Lin, Chaofan and Dong, Chen and Ling, Chenchen and others},
  journal={arXiv preprint arXiv:2606.19348},
  year={2026}
}

@article{jin2026entropy,
  title={Entropy-aware on-policy distillation of language models},
  author={Jin, Woogyeol and Min, Taywon and Yang, Yongjin and Wei, Dennis and Zhou, Yi and Kadhe, Swanand Ravindra and Baracaldo, Nathalie and Lee, Kimin},
  journal={arXiv preprint arXiv:2603.07079},
  year={2026}
}

@inproceedings{gu2024minillm,
  title={Minillm: Knowledge distillation of large language models},
  author={Gu, Yuxian and Dong, Li and Wei, Furu and Huang, Minlie},
  booktitle={International Conference on Learning Representations},
  volume={2024},
  pages={32694--32717},
  year={2024}
}

@article{li2026rethinking,
  title={Rethinking on-policy distillation of large language models: Phenomenology, mechanism, and recipe},
  author={Li, Yaxuan and Zuo, Yuxin and He, Bingxiang and Zhang, Jinqian and Xiao, Chaojun and Qian, Cheng and Yu, Tianyu and Gao, Huan-ang and Yang, Wenkai and Liu, Zhiyuan and others},
  journal={arXiv preprint arXiv:2604.13016},
  year={2026}
}

@article{song2026survey,
  title={A survey of on-policy distillation for large language models},
  author={Song, Mingyang and Zheng, Mao},
  journal={arXiv preprint arXiv:2604.00626},
  year={2026}
}

@article{kullback1951information,
  title={On information and sufficiency},
  author={Kullback, Solomon and Leibler, Richard A},
  journal={The annals of mathematical statistics},
  volume={22},
  number={1},
  pages={79--86},
  year={1951},
  publisher={JSTOR}
}

@misc{deepseekai2025deepseekr1incentivizingreasoningcapability,
      title={DeepSeek-R1: Incentivizing Reasoning Capability in LLMs via Reinforcement Learning}, 
      author={DeepSeek-AI},
      year={2025},
      eprint={2501.12948},
      archivePrefix={arXiv},
      primaryClass={cs.CL},
      url={https://arxiv.org/abs/2501.12948}, 
}

@article{yang2024qwen25mathtechnicalreportmathematical,
  title={Qwen2.5-Math Technical Report: Toward Mathematical Expert Model via Self-Improvement}, 
  author={An Yang and Beichen Zhang and Binyuan Hui and Bofei Gao and Bowen Yu and Chengpeng Li and Dayiheng Liu and Jianhong Tu and Jingren Zhou and Junyang Lin and Keming Lu and Mingfeng Xue and Runji Lin and Tianyu Liu and Xingzhang Ren and Zhenru Zhang},
  journal={arXiv preprint arXiv:2409.12122},
  year={2024}
}

@article{he2025justrl,
  title={JustRL: Scaling a 1.5 B LLM with a Simple RL Recipe},
  author={He, Bingxiang and Qu, Zekai and Liu, Zeyuan and Chen, Yinghao and Zuo, Yuxin and Qian, Cheng and Zhang, Kaiyan and Chen, Weize and Xiao, Chaojun and Cui, Ganqu and others},
  journal={arXiv preprint arXiv:2512.16649},
  year={2025}
}

@misc{yu2025dapoopensourcellmreinforcement,
      title={DAPO: An Open-Source LLM Reinforcement Learning System at Scale}, 
      author={Qiying Yu and Zheng Zhang and Ruofei Zhu and Yufeng Yuan and Xiaochen Zuo and Yu Yue and Weinan Dai and Tiantian Fan and Gaohong Liu and Lingjun Liu and Xin Liu and Haibin Lin and Zhiqi Lin and Bole Ma and Guangming Sheng and Yuxuan Tong and Chi Zhang and Mofan Zhang and Wang Zhang and Hang Zhu and Jinhua Zhu and Jiaze Chen and Jiangjie Chen and Chengyi Wang and Hongli Yu and Yuxuan Song and Xiangpeng Wei and Hao Zhou and Jingjing Liu and Wei-Ying Ma and Ya-Qin Zhang and Lin Yan and Mu Qiao and Yonghui Wu and Mingxuan Wang},
      year={2025},
      eprint={2503.14476},
      archivePrefix={arXiv},
      primaryClass={cs.LG},
      url={https://arxiv.org/abs/2503.14476}, 
}

@misc{pyatkin2025generalizing,
   title={Generalizing Verifiable Instruction Following}, 
   author={Valentina Pyatkin and Saumya Malik and Victoria Graf and Hamish Ivison and Shengyi Huang and Pradeep Dasigi and Nathan Lambert and Hannaneh Hajishirzi},
   year={2025},
   eprint={TODO},
   archivePrefix={arXiv},
   primaryClass={cs.CL}
}

@misc{wang2026madopdbreakingceilingonpolicy,
      title={MAD-OPD: Breaking the Ceiling in On-Policy Distillation via Multi-Agent Debate}, 
      author={Jianze Wang and Ying Liu and Jinlong Chen and Xuchun Hu and Qilong Zhang and Yu Cao and Jun Wang and Hua Yang and Yong Xie and Qianglong Chen},
      year={2026},
      eprint={2605.01347},
      archivePrefix={arXiv},
      primaryClass={cs.CL},
      url={https://arxiv.org/abs/2605.01347}, 
}

@misc{yu2026weaktostrongonpolicydistillation,
      title={Weak-to-Strong On-Policy Distillation}, 
      author={Fangxu Yu and Weijia Xu and Michael Xu and Tianyi Zhou and Zinan Lin},
      year={2026},
      eprint={2607.26246},
      archivePrefix={arXiv},
      primaryClass={cs.LG},
      url={https://arxiv.org/abs/2607.26246}, 
}

@article{fu2026rethinking,
  title={Rethinking on-policy distillation of large language models ii: One training example},
  author={Fu, Zixuan and He, Bingxiang and Zuo, Yuxin and Huang, Haohuan and Zhang, Jinqian and Xiao, Ruhang and Qian, Cheng and Luo, Qinyu and Gao, Huan-ang and Wang, Yudong and others},
  journal={arXiv preprint arXiv:2609.04172},
  year={2026}
}

@inproceedings{kim2016sequence,
  title={Sequence-level knowledge distillation},
  author={Kim, Yoon and Rush, Alexander M},
  booktitle={Proceedings of the 2016 conference on empirical methods in natural language processing},
  pages={1317--1327},
  year={2016}
}

@article{niu2026breaking,
  title={Breaking the Tokenizer Barrier: On-Policy Distillation across Model Families},
  author={Niu, Yifan and Xiao, Han and Liu, Dongyi and Wang, Zelong and Gong, Dihong and Wang, Yasheng and Li, Jia},
  journal={arXiv preprint arXiv:2606.09456},
  year={2026}
}

@article{wang2026not,
  title={Not all disagreement is learnable: Token teachability in on-policy distillation},
  author={Wang, Yuanyi and Lu, Su and Gu, Yanggan and Wang, Pengkai and Yang, Yifan and Yan, Zhaoyi and Xie, Congkai and Wu, Jianmin and Yang, Hongxia},
  journal={arXiv preprint arXiv:2605.26844},
  year={2026}
}

@article{liu2026teacher,
  title={Teacher-guided policy optimization for on-policy reasoning distillation under large policy divergence},
  author={Liu, Xinyu and Jiao, Kechen and Xiao, Chunyang and Zhao, Runsong and Ruan, Junhao and Li, Bei and Liu, Jiahao and Wang, Qifan and Chen, Xin and Wang, Jingang and others},
  journal={arXiv preprint arXiv:2605.13230},
  year={2026}
}

@article{li2026rethinkingr,
  title={Rethinking Reverse KL as Adaptive Entropy Distillation},
  author={Li, Shizhen and Shen, Zhiyu and Lu, Yuyin and Pang, Yunhe and Song, Jielin and Rao, Yanghui and Wang, Fu Lee},
  journal={arXiv preprint arXiv:2608.14685},
  year={2026}
}

@article{ma2026mopd,
  title={Mopd: Multi-teacher on-policy distillation for capability integration in llm post-training},
  author={Ma, Wenhan and Wei, Jianyu and Zhao, Liang and Zhang, Hailin and Xiao, Bangjun and Li, Lei and Yang, Qibin and Gao, Bofei and Wang, Yudong and Li, Rang and others},
  journal={arXiv preprint arXiv:2606.30406},
  year={2026}
}

@article{gao2026open,
  title={Open-MOPD: Diagnosing and Fixing Capability Imbalance in Multi-Teacher On-Policy Distillation},
  author={Gao, Huan-ang and Chi, Haohan and Yan, Yong and Feng, Shiyuan and Wu, Hanlin and Jiang, Zheng and He, Bingxiang and Ma, Wei-Ying and Zhang, Ya-Qin and Zhou, Hao},
  journal={arXiv preprint arXiv:2608.19098},
  year={2026}
}

@article{burns2023weak,
  title={Weak-to-strong generalization: Eliciting strong capabilities with weak supervision},
  author={Burns, Collin and Izmailov, Pavel and Kirchner, Jan Hendrik and Baker, Bowen and Gao, Leo and Aschenbrenner, Leopold and Chen, Yining and Ecoffet, Adrien and Joglekar, Manas and Leike, Jan and others},
  journal={arXiv preprint arXiv:2312.09390},
  year={2023}
}

@inproceedings{yuan2026incentivizing,
  title={Incentivizing strong reasoning from weak supervision},
  author={Yuan, Yige and Xiao, Teng and Tao, Shuchang and Wang, Xue and Gao, Jinyang and Ding, Bolin and Xu, Bingbing},
  booktitle={Proceedings of the 19th Conference of the European Chapter of the Association for Computational Linguistics (Volume 1: Long Papers)},
  pages={7138--7156},
  year={2026}
}

@inproceedings{wang2026drift,
  title={Drift: Learning from abundant user dissatisfaction in real-world preference learning},
  author={Wang, Yifan and Li, Bolian and Wu, Junlin and Tan, Zhaoxuan and Liu, Zheli and Zhang, Ruqi and Grama, Ananth and Zeng, Qingkai},
  booktitle={International Conference on Learning Representations},
  volume={2026},
  pages={120457--120479},
  year={2026}
}

@article{feng2026weak,
  title={Weak-to-Strong Generalization via Direct On-Policy Distillation},
  author={Feng, Shiyuan and Gao, Huan-ang and Chi, Haohan and Wu, Hanlin and Zhang, Zhilong and Jiang, Zheng and He, Bingxiang and Ma, Wei-Ying and Zhang, Ya-Qin and Zhou, Hao},
  journal={arXiv preprint arXiv:2607.05394},
  year={2026}
}

@misc{aime24,
      title={American Invitational Mathematics Examination (AIME) 2024}, 
      author={Zhang, Yifan and Math-AI, Team},
      year={2024},
}

@misc{aime25,
      title={American Invitational Mathematics Examination (AIME) 2025}, 
      author={Zhang, Yifan and Math-AI, Team},
      year={2025},
}

@misc{aime26,
      title={American Invitational Mathematics Examination (AIME) 2026}, 
      author={Zhang, Yifan and Math-AI, Team},
      year={2026},
}

@inproceedings{gao2025omni,
  title={Omni-math: A universal olympiad level mathematic benchmark for large language models},
  author={Gao, Bofei and Song, Feifan and Yang, Zhe and Cai, Zefan and Miao, Yibo and Dong, Qingxiu and Li, Lei and Ma, Chenghao and Chen, Liang and Tang, Zhengyang and others},
  booktitle={International Conference on Learning Representations},
  volume={2025},
  pages={100540--100569},
  year={2025}
}

@article{armandpour2026unmasking,
  title={Unmasking on-policy distillation: Where it helps, where it hurts, and why},
  author={Armandpour, Mohammadreza and Ilhan, Fatih and Harrison, David and Jaiswal, Ajay and Hoang, Duc NM and Faghri, Fartash and Zhang, Yizhe and Cho, Minsik and Farajtabar, Mehrdad},
  journal={arXiv preprint arXiv:2605.10889},
  year={2026}
}

@inproceedings{rajpurkar2018know,
  title={Know what you don’t know: Unanswerable questions for SQuAD},
  author={Rajpurkar, Pranav and Jia, Robin and Liang, Percy},
  booktitle={Proceedings of the 56th Annual Meeting of the Association for Computational Linguistics (Volume 2: Short Papers)},
  pages={784--789},
  year={2018}
}
\bibliographystyle{iclr2027_conference}

\appendix
% Insert after the existing \appendix command, following the references.
% This section requires no additional packages.
\section{Implementation Details}
\label{app:implementation}

Table \ref{tab:hyperparameters} summarizes the training hyperparameters
for mathematical reasoning, instruction following, and question
answering with abstention. For evaluation, we use temperature 0.7
for Pass@16 and temperature 0.5 for Pass@1, with top-$p = 0.95$. For IF evaluation, we exclude prompts with invalid grader outputs for any compared model within each configuration, yielding common evaluation subsets of 6,900 prompts for Qwen3 and 6,906 for Llama from the original 7,000 held-out prompts.

\begin{table}[ht]
    \centering
    \caption{Training hyperparameters across the three tasks.
    Maximum response lengths are measured in tokens.}
    \label{tab:hyperparameters}
    \small
    \begin{tabular}{lccc}
        \toprule
        Hyperparameter & Math (DAPO) & IF & SQuAD 2.0 \\
        \midrule
        $\delta$                      & 0.05 & 0.05 & 0.05\\
        $\tau$                        & 1.0 & 1.0 & 1.0\\
        Optimizer                     & AdamW & AdamW & AdamW \\
        Learning rate                 & $1\times10^{-6}$ & $1\times10^{-6}$ & $1\times10^{-6}$ \\
        Weight decay                  & 0.01 & 0.01 & 0.01 \\
        Training epochs               & 3 & 3 & 3 \\
        Prompt batch size             & 64 & 64 & 64 \\
        Responses per prompt          & 4 & 4 & 4 \\
        PPO mini-batch size            & 64 & 64 & 64 \\
        Training sampling temperature  & 1.0 & 1.0 & 1.0 \\
        Distillation support size      & 16 & 16 & 16 \\
        Maximum response length        & 7,168 & 4,096 & 4,096 \\
        \bottomrule
    \end{tabular}
\end{table}

\section{Ablation Study}
\label{app:ablation}
% Requires \usepackage{booktabs}, \usepackage{xcolor}, and \usepackage{colortbl}.
\begin{table}[t]
    \centering
    \caption{Training data and supervision branch ablations on mathematical reasoning. The filtered
    subset contains 6,016 DAPO-Math-17k prompts, which is consistent with the main experiment. The positive only and
    negative only variants activate only $m_v^{+}$ and $m_v^{-}$, respectively.
    Bold and underlined scores indicate the best and second-best results within each training-data group.}
    \label{tab:math-ablation}
    \vspace{\baselineskip}
    \normalsize
    \setlength{\tabcolsep}{3pt}
    \renewcommand{\arraystretch}{1.00}
    \setlength{\aboverulesep}{1pt}
    \setlength{\belowrulesep}{1pt}
    \begin{tabular*}{\linewidth}{@{\extracolsep{\fill}}l*{6}{c}@{}}
        \toprule
        \textbf{Method}
        & \shortstack[c]{AIME24\\Pass@16 $\uparrow$}
        & \shortstack[c]{AIME25\\Pass@16 $\uparrow$}
        & \shortstack[c]{AIME26\\Pass@16 $\uparrow$}
        & \shortstack[c]{AMC23\\Pass@16 $\uparrow$}
        & \shortstack[c]{Avg.\\Pass@16 $\uparrow$}
        & \shortstack[c]{Omni-Math\\Pass@1 $\uparrow$} \\
        \midrule
        \rowcolor{black!6}
        \multicolumn{7}{c}{\textit{Full DAPO-Math-17k}} \\
        \midrule
        FKL & \underline{76.67} & \textbf{60.00} & \underline{36.67} & 89.16 & 72.83 & 18.54 \\
        RKL & \textbf{80.00} & \textbf{60.00} & 33.33 & \underline{93.98} & \underline{75.14} & 19.32 \\
        JSD & \underline{76.67} & \underline{53.33} & \textbf{43.33} & \underline{93.98} & \underline{75.14} & \underline{19.53} \\
        \midrule
        \textbf{cOPD (ours)} & \textbf{80.00} & \textbf{60.00} & \underline{36.67} & \textbf{95.18} & \textbf{76.30} & \textbf{24.14} \\
        \midrule
        \rowcolor{black!6}
        \multicolumn{7}{c}{\textit{Filtered DAPO-Math-17k subset (6,016 prompts)}} \\
        \midrule
        FKL & 76.67 & 53.33 & \underline{43.33} & 91.57 & 73.99 & 12.19 \\
        RKL & 73.33 & 46.67 & 40.00 & \underline{93.98} & 72.83 & 15.14 \\
        JSD & 76.67 & \underline{56.67} & \underline{43.33} & 92.77 & 75.14 & 14.14 \\
        \midrule
        cOPD ($m_v^{+}$ only) & \textbf{83.33} & \textbf{60.00} & \underline{43.33} & 92.77 & \underline{76.88} & 20.95 \\
        cOPD ($m_v^{-}$ only) & 73.33 & 53.33 & 40.00 & 91.57 & 72.83 & \textbf{26.76} \\
        \textbf{cOPD (ours)} & \underline{80.00} & \underline{56.67} & \textbf{46.67} & \textbf{95.18} & \textbf{77.46} & \underline{23.82} \\
        \bottomrule
    \end{tabular*}
\end{table}

Table \ref{tab:math-ablation} examines training set selection and branch contributions of cOPD.
When trained on the full DAPO-Math-17k dataset, cOPD exceeds the strongest single-teacher baselines in pooled Pass@16 (76.30\% versus 75.14\% for RKL and JSD) and Omni-Math Pass@1 (24.14\% versus 19.53\% for JSD). 
Training on the teacher-separated subset of 6,016 prompts yields 77.46\% pooled Pass@16 and 23.82\% Omni-Math Pass@1, exceeding every full-data single-teacher OPD baseline on both metrics despite using fewer student-training prompts. Full data cOPD remains competitive, indicating that its performance is not restricted to the filtered subset.
Among the cOPD variants, positive only outperforms negative only on all four AIME/AMC Pass@16 benchmarks, whereas negative-only leads on Omni-Math Pass@1. Combining the branches gives the highest pooled Pass@16 among these variants and an intermediate Omni-Math score, balancing their different strengths across the two evaluation metrics.

\section{Examples for instruction following}
\label{app:if_examples}

We show abbreviated responses to a shared prompt.
The cOPD and RKL examples correspond to the highlighted cOPD-only cluster
and circled word limit failure in Figure \ref{fig:if_umap}.
Markdown markers are preserved literally, and \texttt{[...]} marks
omitted text.
\noindent
\begin{tabular}{@{}>{\raggedright\arraybackslash}p{\linewidth}@{}}
\toprule
\textbf{Shared prompt} \\
\midrule
Which avenue in France was renamed after Victor Hugo? Highlight at
least 7 sections in your answer using Markdown, for example,
\texttt{*highlighted section*}. Answer in fewer than 606 words.
The letter v must appear at least 13 times in your response. \\
\bottomrule
\end{tabular}

\begin{table}[!htbp]
\centering
\caption{cOPD response (all instruction constraints satisfied).}
\label{tab:if-example-copd}
\begin{tabular}{@{}>{\raggedright\arraybackslash}p{\linewidth}@{}}
\toprule
The avenue in France, \texttt{*Avenue Victor Hugo*}, was renamed after
the famous writer Victor Hugo. \texttt{[...]}
\texttt{*V*} is an important element in the avenue's name, reflecting
the honor paid to him.
\par\medskip
\texttt{*V*} also appears in the names of many other roads in Paris,
such as \texttt{*V\'etheuil*}, \texttt{*Vieux-Paris*}, and
\texttt{*Ville-Victor*}. \texttt{[...]} \\
% \midrule
% \textit{Note.} The full response also contains repetition, but remains
% within the word limit. \\
\bottomrule
\end{tabular}
\end{table}

\begin{table}[!htbp]
\centering
\caption{RKL response (word-limit violation).}
\label{tab:if-example-rkl}
\begin{tabular}{@{}>{\raggedright\arraybackslash}p{\linewidth}@{}}
\toprule
The avenue in France renamed after Victor Hugo is
\texttt{**Avenue Victor Hugo**}. \texttt{[...]}
\par\medskip
With Victor Hugo's presence, this avenue has become a symbol of
French culture and history. \texttt{*highlighted section*} \texttt{[...]}
With Victor Hugo's presence, this avenue has become a symbol of
French culture and history. \texttt{*highlighted section*} \texttt{[...]}
With the participation of many artists and writers, Avenue Victor Hugo has \\
\midrule
\textit{Note.} The full recorded response contains extensive repetition
and ends mid-sentence, as preserved above. \\
\bottomrule
\end{tabular}
\end{table}

\begin{table}[!htbp]
\centering
\caption{MAD-OPD response.}
\label{tab:if-example-mad}
\begin{tabular}{@{}>{\raggedright\arraybackslash}p{\linewidth}@{}}
\toprule
The avenue in France renamed after Victor Hugo is
\texttt{**Avenue Victor Hugo**}. \texttt{[...]}
\par\medskip
\texttt{*highlighted section*}\par
Avenue Victor Hugo also passes through \texttt{**Rue de Rivoli Street**},
\texttt{**Rue de la Pompe Street**}, and \texttt{**Rue de la Pompe Street**}.
\texttt{[...]} \\
\bottomrule
\end{tabular}
\end{table}

\begin{table}[!htbp]
\centering
\caption{W2S-OPD response.}
\label{tab:if-example-w2s}
\begin{tabular}{@{}>{\raggedright\arraybackslash}p{\linewidth}@{}}
\toprule
The avenue in France renamed after Victor Hugo is
\texttt{**Avenue Victor Hugo**}. \texttt{[...]}
\par\medskip
\texttt{*highlighted section*}\par
The road also hosts many cultural and artistic events, such as art
exhibitions and performances that take place regularly. \texttt{[...]}
With 13 occurrences of the letter \texttt{**v**}, Avenue Victor Hugo
is one of the most important roads in Paris, closely associated with
French culture and history. \\
\bottomrule
\end{tabular}
\end{table}
\FloatBarrier

% \section{Appendix}

\end{document}